\documentclass[10pt]{article} 

\usepackage[preprint]{rlc_new}

\usepackage{amssymb}            
\usepackage{mathtools}          
\usepackage{mathrsfs}           
\usepackage{graphicx}           
\usepackage{subcaption}         
\usepackage[space]{grffile}     
\usepackage{url}                

\usepackage{algorithm}

\usepackage{natbib}
\usepackage{eso-pic} 
\usepackage{forloop}
\usepackage{url}

\usepackage{microtype}
\usepackage{graphicx}
\usepackage{booktabs} 

\usepackage{hyperref}

\usepackage{amsmath}
\usepackage{amssymb}
\usepackage{mathtools}
\usepackage{amsthm}
\usepackage{xspace}

\theoremstyle{plain}

\theoremstyle{definition}

\theoremstyle{remark}

\usepackage{bm}
\usepackage{algpseudocode}
\usepackage{caption}
\usepackage{subcaption}
\usepackage{multicol}
\usepackage{multirow}
\usepackage{soul}
\usepackage{siunitx}
\usepackage{wrapfig}

\newcommand{\STAB}[1]{\begin{tabular}{@{}c@{}}#1\end{tabular}}
\newcolumntype{N}{r@{\hspace{.7mm}}c@{\hspace{.7mm}}l}

\newcommand{\ptarg}{p_{\mkern1mu\text{target}}}
\newcommand{\qtarg}{q_{\mkern1mu\text{target}}}
\newcommand{\poff}{p_{\mkern1mu\text{off}}}
\newcommand{\qoff}{q_{\mkern1mu\text{off}}}
\newcommand{\doff}{\mathcal{D}_{\text{off}}}

\definecolor{tab_blue}{rgb}{0.0, 0.5, 1.0}
\definecolor{tab_green}{rgb}{0.4, 0.69, 0.2}
\definecolor{pgd}{RGB}{245, 115, 0}
\definecolor{pgd_comp}{RGB}{175, 255, 255}
\sethlcolor{pgd_comp}
\newcommand{\mhl}[1]{\colorbox{pgd_comp}{$\displaystyle #1$}\hspace{-1mm}}

\title{Policy-Guided Diffusion}

\author{
}

\begin{document}

\maketitle

\renewcommand{\thefootnote}{\fnsymbol{footnote}}

\vspace{-10mm}
\centerline{\textbf{
Matthew T.\ Jackson\footnote{Correspondence to \href{mailto:jackson@robots.ox.ac.uk}{jackson@robots.ox.ac.uk}.}\hspace{3mm}
Michael T.\ Matthews\hspace{3mm}
Cong Lu\hspace{3mm}
Benjamin Ellis
}}
\centerline{\textbf{
Shimon Whiteson\footnotemark{}\hspace{3mm}
Jakob N.\ Foerster\footnotemark[2]{} 
}}
\footnotetext{Equal supervision.}
\vspace{2mm}
\centerline{University of Oxford}

\vspace{10mm}

\begin{abstract}
In many real-world settings, agents must learn from an \textit{offline} dataset gathered by some prior behavior policy. 
Such a setting naturally leads to distribution shift between the behavior policy and the target policy being trained---requiring policy conservatism to avoid instability and overestimation bias.
Autoregressive world models offer a different solution to this by generating synthetic, on-policy experience.
However, in practice, model rollouts must be severely truncated to avoid compounding error.
As an alternative, we propose \textit{policy-guided diffusion}.
Our method uses diffusion models to generate entire trajectories under the behavior distribution, applying \textit{guidance} from the target policy to move synthetic experience further on-policy.
We show that policy-guided diffusion models a regularized form of the target distribution that balances action likelihood under both the target and behavior policies, leading to plausible trajectories with high target policy probability, 
while retaining a lower dynamics error than an offline world model baseline.
Using synthetic experience from policy-guided diffusion as a drop-in substitute for real data, we demonstrate significant improvements in performance across a range of standard offline reinforcement learning algorithms and environments.
Our approach provides an effective alternative to autoregressive offline world models, opening the door to the controllable generation of synthetic training data.
\end{abstract}

\section{Introduction}
\label{intro}
A key obstacle to the real-world adoption of reinforcement learning~\citep[RL,][]{Sutton1998} is its notorious sample inefficiency, preventing agents from being trained on environments with expensive or slow online data collection.
A closely related challenge arises in environments where exploration, required by standard RL methods, is inherently dangerous, limiting their applicability.
Yet many such settings come with an abundance of pre-collected or offline experience, gathered under one or more \textit{behavior} policies~\citep{mopo}.
These settings enable the application of offline RL~\citep{levine2020offline}, where a policy is optimized from an offline dataset without access to the environment.
However, the distribution shift between the \textit{target} policy (i.e., the policy being optimized) and the collected data poses many challenges~\citep{kumar2020conservative, kostrikov2021offline, bcq}.

In particular, \textit{distribution shift} between the target and behavior policies leads to an out-of-sample issue: since the goal of offline RL is to \textit{exceed} the performance of the behavior policy, the distribution of state-action pairs sampled by the target policy necessarily differs from that of the behavior policy, and its samples are therefore underrepresented (or unavailable) in the offline dataset.
However, the maximizing nature of RL classically leads to overestimation bias when generalizing to rarely seen state-action pairs, resulting in an overly optimistic target policy.
As a solution, most previous \textit{model-free} work has proposed severe regularization of the target policy---such as penalizing value estimates in uncertain states~\citep{kumar2020conservative, an2021uncertainty} or regularizing it towards the behavior policy~\citep{fujimoto2021minimalist}---sacrificing target policy performance for stability.

\begin{figure}[t!]
\centering
\includegraphics[width=\textwidth]{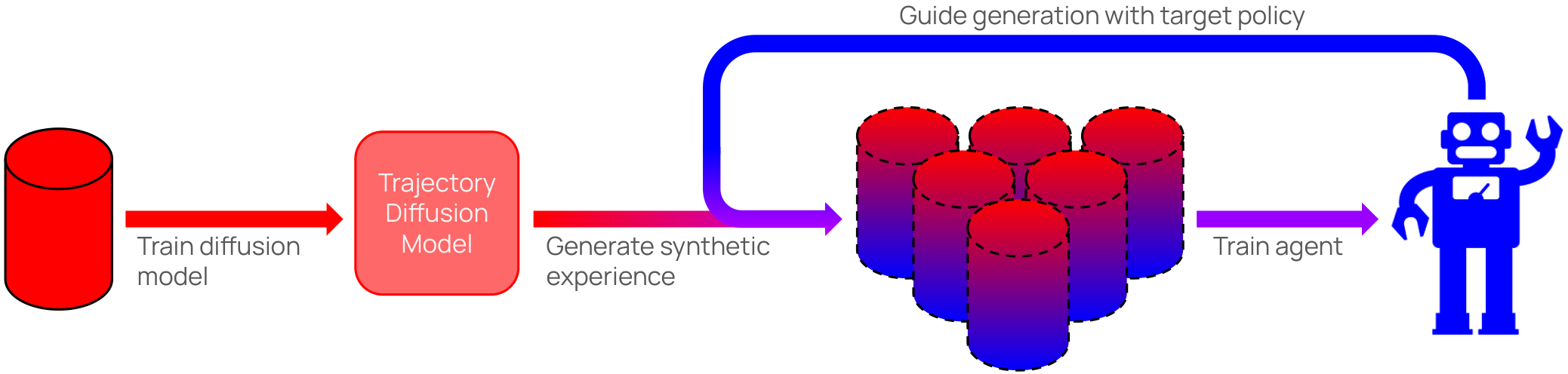}
\vspace{-4mm}
\caption{
Offline reinforcement learning with policy-guided diffusion. Offline data from a \textcolor{red}{behavior policy} is first used to train a trajectory diffusion model. Synthetic experience is then generated with diffusion, guided by the \textcolor{blue}{target policy} in order to move trajectories further on-policy. An agent is then trained for multiple steps on the synthetic dataset, before it is regenerated.
}
\vspace{-4mm}
\label{fig:overview}
\end{figure}

In this paper, we focus on an alternative class of methods: generating \textit{synthetic experience} to both augment the offline dataset and lessen the out-of-sample issue.
Prior methods in this area use a \textit{model-based} approach~\citep[see \autoref{sec:autoregressive}]{mopo, kidambi2020morel, augwm}, in which a single-step world model is learned from the offline dataset, which the target policy interacts with to generate synthetic on-policy training experience.
While this allows the target policy to sample synthetic trajectories under its own action distribution, compounding model error usually forces these methods to severely truncate model rollouts to a handful of interactions.
Thus, there are two options which trade off coverage and bias.
The first is to roll out from the initial state, which is unbiased but lacks coverage. 
The second is to roll out from states randomly sampled from the data set, which increases coverage but introduces bias.
Neither option fully addresses the difference in observed states between the behavior and target policy when deployed, nor the out-of-sample issue mentioned above.

\begin{figure}[t!]
\centering
\vspace{-2mm}
\includegraphics[width=\textwidth]{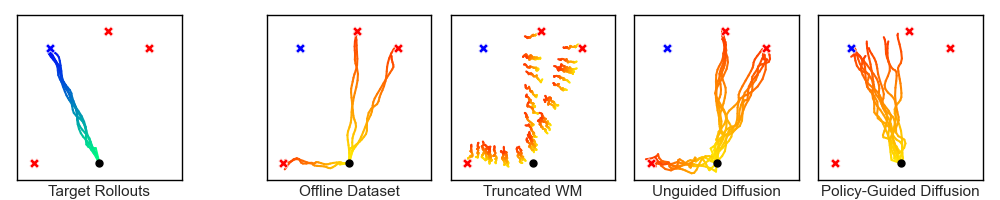}
\vspace{-4mm}
\caption{
Trajectories from an illustrative 2D environment, in which the start location is indicated by $\bullet$ and the goals for the behavior and target policies are indicated by \textcolor{red}{$\mathbf{\times}$} and \textcolor{blue}{$\mathbf{\times}$}.
\textbf{Left:} Rollouts from the \textcolor{blue}{target policy} in the real environment.
\textbf{Right:} 
Offline datasets gathered by the \textcolor{red}{behavior policy} suffer from distribution shift and limited sample size.
Truncated world models~\citep{mopo, kidambi2020morel} previously used in offline model-based reinforcement learning offer a partial solution to this problem but suffer from bias due to short rollouts.
Meanwhile, unguided diffusion~\citep{lu2023synthetic} can increase the sample size, but maintains the original distribution shift.
In contrast, policy-guided diffusion samples from a regularized target distribution, generating entire trajectories with low transition error but higher likelihood under the target distribution.
}
\label{fig:teaser}
\end{figure}

Instead, we propose \textit{policy-guided diffusion} (PGD, \autoref{fig:overview}), which avoids compounding error by modeling entire trajectories (\autoref{sec:joint}) rather than single-step transitions.
To achieve this, we train a \textit{diffusion model} on the offline dataset, from which we can sample synthetic trajectories under the behavior policy.
However, while this addresses \textit{data sparsity}, these trajectories are still \textit{off-distribution} from our target policy.
Therefore, inspired by classifier-guided diffusion~\citep{dhariwal2021diffusion}, we apply \textit{guidance} from the target policy to shift the sampling distribution towards that of the target policy.
At each diffusion step, our guidance term directly increases the likelihood of sampled synthetic actions under the target policy, while the diffusion model updates the entire trajectory towards those in the dataset.
This yields a regularized target distribution that we name the \emph{behavior-regularized target distribution}, ensuring actions do not diverge too far from the behavior policy, limiting generalization error.
As a result, PGD does not suffer from compounding error, while also generating synthetic trajectories that are more representative of the target policy.
We illustrate this point in \autoref{fig:teaser}.

Our approach results in consistent improvements in offline RL performance for agents trained on policy-guided synthetic data, compared to those trained on unguided synthetic or real data.
We evaluate using the standard TD3+BC~\citep{fujimoto2021minimalist} and IQL~\citep{kostrikov2021offline} algorithms across a variety of D4RL~\citep{fu2020d4rl} datasets.
Notably, we see a statistically significant 11.2\%  improvement in performance for the TD3+BC algorithm aggregated across MuJoCo~\citep{mujoco} locomotion tasks compared to training on the real data, with no algorithmic changes.
Our results also extend to even larger improvements for the challenging Maze2d navigation environments.
Furthermore, we analyze synthetic trajectories generated by PGD and show that PGD achieves lower dynamics error than PETS~\citep{chua2018deep}---a prior offline model-based method---while matching the target policy likelihood of PETS.
Together, our experiments illustrate the potential of PGD as an effective drop-in replacement for real data---across agents, environments, and behavior policies.

\section{Background}
\label{background}

\subsection{Offline Reinforcement Learning}
\label{sec:offline_rl}
\paragraph{Formulation}
We adopt the standard reinforcement learning formulation, in which an agent acts in a Markov Decision Process~\citep[MDP,][]{Sutton1998}. An MDP is defined as a tuple $M = \langle \mathcal{S}, \mathcal{A}, p_0, T, R, H \rangle$, where $s\in\mathcal{S}$ and $a\in\mathcal{A}$ are the state and action spaces, $p_0(s_0)$ is a probability distribution over the initial state, $T(s_{t+1}|s_t,a_t)$ is a conditional probability distribution defining the transition dynamics, $R: \mathcal{S} \times \mathcal{A} \xrightarrow{} \mathbb{R}$ is the reward function, $\gamma$ is the discount factor, and $H$ is the horizon.

In RL, we learn a policy $\pi(a|s)$ that defines a conditional probability distribution over actions for each state, inducing a distribution over trajectories $\bm{\tau} \coloneqq (s_0, a_1, r_1, s_1,\ldots, s_H)$ given by
\begin{equation}
    \label{eq:autoregressive}
    p_{\pi, M}(\bm{\tau}) = p_0(s_0) \prod_{t=0}^{H-1} \pi(a_t|s_t) \cdot T(s_{t+1}|s_t, a_t)\text{,}
\end{equation}
omitting the reward function throughout our work for conciseness.
Our goal is to learn a policy that maximizes the expected return, defined as $\mathbb{E}_{p_{\pi, M}}[V(\bm{\tau})]$ where $V(\bm{\tau}) \coloneqq \sum_{t=0}^{H}r_{t}$ is the return of a trajectory.
The \textit{offline RL} setting~\citep{levine2020offline} extends this, preventing the agent from interacting with the environment and instead presenting it with a dataset of trajectories $\bm{\tau} \in \doff$ gathered by some unknown behavior policy $\pi_{\text{off}}$, with which to optimize a target policy $\pi_{\text{target}}$.

\paragraph{Out-of-Sample Generalization}
The core challenge of offline RL emerges from the distribution shift between the \emph{behavior distribution} $p_{\pi_\text{off}, M}(\bm{\tau})$ and the \emph{target distribution} $p_{\pi_\text{target}, M}(\bm{\tau})$, which are otherwise denoted $\poff(\bm{\tau})$ and $\ptarg(\bm{\tau})$ for conciseness.
Optimization of $\pi_{\text{target}}$ on $\doff$ can lead to catastrophic value overestimation at unobserved actions, a problem termed the \emph{out-of-sample issue}~\citep{kostrikov2021offline}.
As such, model-free offline algorithms typically regularize the policy towards the behavior distribution, either explicitly~\citep{fujimoto2021minimalist, kumar2020conservative} or implicitly~\citep{kostrikov2021offline}.

Alternatively, prior work proposes learning a single-step world model $\mathcal{M}$ from $\doff$~\citep{mopo, kidambi2020morel, lu2022revisiting}.
By rolling out the target policy using $\mathcal{M}$, we generate trajectories $\bm{\tau} \sim \ptarg(\bm{\tau})$, with the aim of avoiding distribution shift.
However, in practice, this technique only pushes the generalization issue into the world model.
In particular, RL policies are prone to exploiting errors in the world model, which can compound over the course of an episode.
When combined with typical maximizing operations used in off-policy RL, this results in value overestimation bias~\citep{sims2024edgeofreach}.

\subsection{Diffusion Models}
\label{sec:diffusion-models}

\paragraph{Definition}
To generate synthetic data, we consider diffusion models~\citep{pmlr-v37-sohl-dickstein15, ddpm}, a class of generative model that allows one to sample from a distribution $p(x)$ by iteratively reversing a forward noising process.
\citet{karras2022elucidating} present an ODE formulation of diffusion which, given a noise schedule $\sigma(i)$ indexed by $i$, mutates data according to
\begin{equation}
    \label{eq:edm_ode} d\bm{x} = -\dot{\sigma}(i)\sigma(i)\nabla_{\bm{x}} \log p\left(\bm{x}; \sigma(i) \right) di\text{,}
\end{equation}
where \mbox{$\dot{\sigma} = \frac{\mathrm{d}\sigma}{\mathrm{d}i}$} and $\nabla_{\bm{x}} \log p\left(\bm{x}; \sigma(i)\right)$ is the score function~\citep{hyvarinen2005estimation}, which points towards areas of high data density.
Intuitively, infinitesimal forward or backward steps of this ODE respectively nudge a sample away from or towards the data.
To generate a sample, we start with pure noise at the highest noise level $\sigma_{\textrm{max}}$, and iteratively denoise in discrete timesteps under \autoref{eq:edm_ode}.

\paragraph{Classifier Guidance}
Our method is designed to augment the data-generating process towards on-policy trajectories from the target distribution $\ptarg(\bm{\tau})$, rather than the behavior distribution $\poff(\bm{\tau})$.
To achieve this, we take inspiration from classifier guidance~\citep{dhariwal2021diffusion}, which leverages a differentiable classifier to augment the score function of a pre-trained diffusion model towards a class-conditional distribution $p(\bm{x}|y)$.
Concretely, this adds a classifier gradient to the score function, giving
\begin{equation}
    \nabla_{\bm{x}}\log p_{\lambda}\left(\bm{x} | y; \sigma(i) \right) = \nabla_{\bm{x}}\log p\left(\bm{x}; \sigma(i) \right) + \lambda \nabla_{\bm{x}}\log p_\theta\left(y | \bm{x} ; \sigma(i) \right) \text{,}
\end{equation}
where $\nabla_{\bm{x}}\log p_\theta\left(y | \bm{x} ; \sigma(i) \right)$ is the gradient of the classifier and $\lambda$ is the guidance weight.

\section{On-Policy Sampling from Offline Data}

Generating synthetic agent experience is a promising approach to solving out-of-sample generalization in offline RL.
By generating experience that is unseen in the dataset, the policy may be directly optimized on OOD samples, thereby moving the generalization problem from the policy to the generative model.
Some prior work has suggested learning a model from the offline dataset~\citep{lu2023synthetic}, thereby sampling synthetic experience from the behavior distribution.
While this improves sample coverage, the approach retains many of the original challenges of offline RL.
As with the behavior policy, the synthetic trajectories may be suboptimal, meaning we still require conservative off-policy RL techniques to train the agent.

Instead, we seek to extend this approach by making our generative model sample from the target distribution.
This reduces the need for conservatism and generates synthetic trajectories with increasing performance as the agent improves over training.
Practically, the effectiveness of this approach depends on how we parameterize each of the terms of the trajectory distribution (\autoref{eq:autoregressive}).
In this section, we consider two parameterizations: autoregressive and direct.

\subsection{Autoregressive Generation --- Model $\bm{\mathcal{T}}$, Sample $\bm{p(s_0)}$}
\label{sec:autoregressive}
The autoregressive---or model-based---approach to generating on-policy data is to use the offline dataset to train a one-step transition model $\mathcal{T}(s_{t+1}|s_t, a_t;\theta)$.
To generate unbiased sample trajectories from the target distribution, we first sample an initial state (i.e., one that starts an episode) from the offline dataset $s_0\sim\doff$.
Next, we roll out our agent in the learned model by iteratively sampling actions from the target policy and approximating environment transitions with the learned dynamics model.
However, compounding error from the transition model usually requires agent rollouts to be much shorter than the environment horizon---such that the agent takes $k\ll H$ steps.\footnote{Typically $k\leq 5$~\citep{janner2019mbpo, mopo}.}
Consequently, any states more than $k$ steps away from any initial state cannot be generated in this manner, limiting the applicability of this approach.

As an approximation, autoregressive methods typically sample initial states from any timestep $s_t\sim\doff$ in the offline dataset.
Given a truncated rollout length $k$, this may be seen as approximating the sub-trajectory distribution---i.e., the trajectory from time $t$ to $t+k$---given by
\begin{align}
    \ptarg(\bm{\tau}_{t:t+k};\theta) &= \ptarg(s_t)\cdot \ptarg(\bm{\tau}_{t:t+k} | s_t;\theta),
\end{align}
by instead modeling
\begin{align}
    \mathcal{F}(\bm{\tau}_{t:t+k};\theta) &= \poff(s_t)\cdot \ptarg(\bm{\tau}_{t:t+k} | s_t;\theta) \label{eq:trunc-wm} \text{.}
\end{align}
Here, we denote the stationary state distributions of the target and behavior policies at time $t$ by $\ptarg(s_t)$ and $\poff(s_t)$ respectively, and define the conditional sub-trajectory distribution as
\begin{align}
    \label{eq:marginal-ar}
    \ptarg(\bm{\tau}_{t:t+k} | s_t;\theta) \coloneqq \prod_{j=0}^{k-1} &\pi_{\text{target}}(a_{t+j}|s_{t+j}) \cdot \mathcal{T}(s_{t+j+1}|s_{t+j}, a_{t+j};\theta)\text{.}
\end{align}

When generating trajectories from this distribution, the difference between $\ptarg(s_t)$ and $\poff(s_t)$ biases the start of rollouts towards states visited by the behavior policy.
Furthermore, we still require $k$ to be small to avoid compounding error.
In combination, sampling from the offline dataset ``anchors'' synthetic rollouts to states in the offline dataset, while truncated rollouts prevent synthetic trajectories from moving far from this anchor.
Therefore, the practical application of autoregressive generation leads to a strong bias towards the behavior distribution and fails to address the out-of-sample problem.

\subsection{Direct Generation --- Model $\bm{p_{\mkern1mu\text{\textbf{off}}}(\tau)}$}
\label{sec:joint}
As an alternative to autoregressive generation, we can parameterize the target distribution by \textit{directly modeling} the behavior distribution, as follows:
\begin{align}
    \notag \ptarg(\bm{\tau}) &= p(s_0) \prod_{t=0}^{H-1} \pi_{\text{target}}(a_t|s_t) \cdot \mathcal{T}(s_{t+1}|s_t, a_t) \\
    \notag &= p(s_0) \prod_{t=0}^{H-1} \frac{\pi_{\text{target}}(a_t|s_t)}{\pi_{\text{off}}(a_t|s_t)} \cdot \pi_{\text{off}}(a_t|s_t) \cdot \mathcal{T}(s_{t+1}|s_t, a_t) \\
    \notag &= \poff(\bm{\tau}) \prod_{t=0}^{H-1} w_{a_t, s_t} \\
    & \approx \poff(\bm{\tau};\theta) \prod_{t=0}^{H-1} w_{a_t, s_t} = \ptarg(\bm{\tau};\theta) \label{eq:importance}
\end{align}
where $w_{a, s} \coloneqq \frac{\pi_{\text{target}}(a|s)}{\pi_{\text{off}}(a|s)}$ denotes the importance sampling weight for $(a, s)$~\citep{precup2000eligibility}.
This directly parameterizes the behavior distribution $\poff(\bm{\tau};\theta)$---which may be learned by modeling entire trajectories on the offline dataset---and adjusts their likelihoods by the relative probabilities of actions $w_{a_t, s_t}$ under the target and behavior policies.
By jointly modeling the initial state distribution, transition function, and behavior policy, such a parameterization is not required to enforce the Markov property.
As a result, it can directly generate entire trajectories, thereby \textit{avoiding the compounding model error} suffered by autoregressive methods when iteratively generating transitions.

However, computing $w_{a_t, s_t}$ requires access to the behavior policy $\pi_{\text{off}}(a|s)$, which is not assumed in offline RL.
Prior work has explored modeling the behavior policy from the offline dataset and using this to compute importance sampling corrections.
However, products of many importance weights can lead to problems with high variance~\citep{10.5555/645529.658134, levine2020offline}.

\section{Policy-Guided Diffusion\label{sec:method}}

In this work, we propose a method following the direct generation approach outlined in \autoref{sec:joint}, named \textit{policy-guided diffusion} (PGD, \autoref{alg:generation}).
Following the success of diffusion models at generating trajectories~\citep{janner2022diffuser, lu2023synthetic}, we first train a trajectory-level diffusion model on the offline dataset to model the behavior distribution.
Then, inspired by classifier-guided diffusion (\autoref{sec:diffusion-models}), we guide the diffusion process using the target policy to move closer to the target distribution.
Specifically, during the denoising process, we compute the gradient of the action distribution for each action under the target policy, using it to augment the diffusion process towards high-probability actions.
In doing so, we approximate a regularized target distribution that equally weights action likelihoods under the behavior and target policies.

In this section, we derive PGD as an approximation of the behavior-regularized target distribution (\autoref{sec:equi-policy}), then describe practical details for controlling and stabilizing policy guidance (\autoref{sec:stable-guidance}). We provide a summary of PGD against alternative sources of training data in \autoref{tab:methods}.

\begin{algorithm}[htb]
    \flushleft
    \caption{Trajectory sampling via {\color{pgd}policy-guided} diffusion --- based on \citet{karras2022elucidating}.}
    \label{alg:generation}
    \begin{algorithmic}[1]
        \State {\bfseries Parameters:} Noise schedule $\sigma_n$, guidance schedule $\lambda_n$, noise factor $\gamma_n$, noise level $S_{\text{noise}}$, diffusion steps $N_{\text{diffusion}}$
        \State {\bfseries Required:} Denoiser model $D_{\theta}$, target policy $\pi_{\phi}$
        \State {\bfseries sample} $\bm{\tau}_0 \sim \mathcal{N}(\bm{0}, \sigma_0^2\bm{I})$ \Comment{Sample random noise trajectory}
        \For{$n=0$ {\bfseries to} $N_{\text{diffusion}}-1$}
            \State {\bfseries sample} $\epsilon_n \sim \mathcal{N}(\bm{0}, S^2_{\text{noise}}\bm{I})$ \Comment{Temporarily increase noise level}
            \State $\hat{\sigma}_n \leftarrow \sigma_n + \gamma_n \sigma_n$
            \State $\bm{\hat{\tau}}_n \leftarrow \bm{\tau}_n + \sqrt{\hat{\sigma}_n^2 - \sigma_n^2}\bm{\epsilon}_n$
            \State $\bm{\bar{\tau}}_n \leftarrow D_{\theta}(\bm{\hat{\tau}}_n;\hat{\sigma}_n)$ \Comment{Estimate denoised trajectory}
            \State $\bm{d}_n \leftarrow \left(\bm{\hat{\tau}}_n - \bm{\bar{\tau}}_n\right)/\hat{\sigma}_n$ \Comment{Evaluate $\frac{\partial \bm{\tau}}{\partial \sigma}$ at $\hat{\sigma}_n$} \color{pgd}
            \State $\bm{g}_n \leftarrow \nabla_{\bar{\tau}_n^{\text{actions}}} \pi_{\phi}(\bm{\bar{\tau}}_n^{\text{actions}} | \bm{\bar{\tau}}_n^{\text{states}})$ \Comment{Compute denoised action gradient}
            \State $\bm{\hat{\tau}}_n^{\text{actions}} \leftarrow \bm{\hat{\tau}}_n^{\text{actions}} + \lambda_n (\bm{g}_n /\lVert\bm{g}_n\rVert_2)$ \Comment{Apply policy guidance to noised actions} \hspace{-1mm}\color{black}
            \State $\bm{\tau}_{n+1} \leftarrow \bm{\hat{\tau}}_n + (\sigma_{n+1} - \hat{\sigma}_n)\bm{d}_n$ \Comment{Apply Euler step}
            \If{$\sigma_{n+1} \neq 0$}
                \State $\bm{d}_n' \leftarrow \left(\bm{\tau}_{n+1} - D_{\theta}(\bm{\tau}_{n+1};\sigma_{n+1})\right)/\sigma_{n+1}$ \Comment {Apply $2^{\text{nd}}$ order correction}
                \State $\bm{\hat{\tau}}_{n+1} \leftarrow \bm{\hat{\tau}}_n + (\sigma_{n+1} - \hat{\sigma}_n)\left(\frac{1}{2}\bm{d}_n + \frac{1}{2}\bm{d}_n'\right)$
            \EndIf
        \EndFor
        \State {\bfseries return} $\bm{\tau}_N$
    \end{algorithmic}
\end{algorithm}

\subsection{\label{sec:equi-policy}Behavior-Regularized Target Distribution}

\paragraph{Policy Guidance Derivation}
To sample a trajectory via diffusion, we require a \textit{noise-conditioned} score function $\nabla_{\bm{\hat{\tau}}} \log p(\bm{\hat{\tau}};\sigma)$ for a noised trajectory $\bm{\hat{\tau}} \coloneqq (\hat{s}_0, \hat{a}_1, \hat{r}_1, \hat{s}_1,\ldots, \hat{s}_H)$ under a distribution $p(\bm{\tau})$ at a noise level $\sigma$.
Given an offline dataset $\doff$, it is straightforward to learn this function under the behavior distribution, $\nabla_{\bm{\hat{\tau}}} \log \poff(\bm{\hat{\tau}};\sigma)$, by training a denoiser model to reconstruct noised trajectories from $\doff$.
However, there is no apparent method to directly model the noise-conditioned score function $\nabla_{\bm{\hat{\tau}}} \log \ptarg(\bm{\hat{\tau}};\sigma)$ for the target distribution (see \autoref{sec:noised-target} for further discussion), meaning we require an approximation.

To achieve this, we consider the score function of a \textit{noise-free} trajectory $\bm{\tau}$ under the target distribution, based on the formulation from \autoref{eq:importance},
\begin{equation}
    \nabla_{\bm{\tau}} \log \ptarg(\bm{\tau}) = \nabla_{\bm{\tau}}\log \poff(\bm{\tau}) + \sum_{t=0}^{H-1} \left(\nabla_{\bm{\tau}}\log \pi_{\text{target}}\left(a_t|s_t\right) - \nabla_{\bm{\tau}}\log \pi_{\text{off}}\left(a_t|s_t\right) \right)\text{.} \label{eq:target-denoise-score}
\end{equation}
In the limit of noise $\sigma \to 0$, the noise-conditioned score function $\nabla_{\bm{\hat{\tau}}} \log \ptarg(\bm{\hat{\tau}};\sigma)$ clearly approaches $\nabla_{\bm{\tau}} \log \ptarg(\bm{\tau})$. Therefore, we may approximate this function by
\begin{equation}
    \nabla_{\bm{\hat{\tau}}} \log \ptarg(\bm{\hat{\tau}};\sigma) \approx
    \nabla_{\bm{\hat{\tau}}}\log \poff(\bm{\hat{\tau}};\sigma) + \sum_{t=0}^{H-1} \left(\nabla_{\bm{\hat{\tau}}}\log \pi_{\text{target}}\left(\hat{a}_t|\hat{s}_t\right) - \nabla_{\bm{\hat{\tau}}}\log \pi_{\text{off}}\left(\hat{a}_t|\hat{s}_t\right) \right)\text{,} \label{eq:target-score}
\end{equation}
for $\sigma \approx 0$. Whilst iteratively denoising under this function (\autoref{sec:diffusion-models}) does not model $\ptarg(\bm{\tau})$ exactly, the score function approaches $\nabla_{\bm{\hat{\tau}}} \log \ptarg(\bm{\hat{\tau}};\sigma)$ towards the end of the denoising process, which we believe provides an effective approximation.

\paragraph{Excluding Behavior Policy Guidance}
As discussed, we may directly model the first term of \autoref{eq:target-score} by training a denoiser model. Furthermore, we may directly compute target policy guidance $\nabla_{\bm{\hat{\tau}}}\log \pi_{\text{target}}\left(\hat{a}_t|\hat{s}_t\right)$---the second term of this approximation---as we assume access to a (differentiable) target policy in the offline RL setting. However, we generally do not have access to the behavior policy, preventing us from computing $\nabla_{\bm{\hat{\tau}}}\log \pi_{\text{off}}\left(\hat{a}_t|\hat{s}_t\right)$. Due to this, we exclude behavior policy guidance from our approximation, resulting in the score function $\nabla_{\bm{\hat{\tau}}}\log \poff(\bm{\hat{\tau}};\sigma) + \sum_{t=0}^{H-1} \nabla_{\bm{\hat{\tau}}}\log \pi_{\text{target}}\left(\hat{a}_t|\hat{s}_t\right)$.
As $\sigma \to 0$, this approaches the score function for a proxy distribution of the form
\begin{align}
    \notag \mathcal{F}(\bm{\tau} ; \pi_{\text{target}}) 
    &\propto \poff(\bm{\tau}) \prod_{t=0}^{H-1} \pi_{\text{target}}(a_t|s_t) \\
    &= \poff(\bm{\tau}) \cdot \qtarg(\bm{\tau}) = \ptarg(\bm{\tau}) \cdot \qoff(\bm{\tau}) \text{,} \label{eq:pgd}
\end{align}
where $\qtarg(\bm{\tau}) \coloneqq \prod_{t=0}^{H-1} \pi_{\text{target}}(a_t|s_t)$ denotes the product of action probabilities under the target policy and $\qoff(\bm{\tau})$ denotes the same quantity under the behavior policy.
Therefore, we hypothesize that excluding behavior policy guidance is an effective form of regularization, as it biases trajectories towards the support of the offline data, thereby limiting model error and the out-of-sample problem.
We refer to $\mathcal{F}(\bm{\tau} ; \pi_{\text{target}})$ as the \textit{behavior-regularized target distribution} due to it balancing action likelihoods under the behavior and target policies, and provide further discussion in \autoref{sec:dist-motiv}.
Finally, as a promising avenue for future work, we note that the behavior policy may be modeled by applying behavior cloning to $\doff$, allowing for the inclusion of behavior policy guidance in the offline RL setting.

\paragraph{Excluding State Guidance}
Target policy guidance $\nabla_{\bm{\hat{\tau}}}\log \pi_{\text{target}}\left(\hat{a}_t|\hat{s}_t\right)$ has non-zero gradients for the state and action at timestep $t$.
In practice, the action component $\nabla_{\hat{a}_t} \log \pi_{\text{target}}\left(\hat{a}_t|\hat{s}_t\right)$ typically has an efficient, closed-form solution, with $\pi_{\text{target}}\left(\hat{a}_t|\hat{s}_t\right)$ commonly being Gaussian for continuous action spaces. In contrast, for neural network policies, the state component $\nabla_{\hat{s}_t} \log \pi_{\text{target}}\left(\hat{a}_t|\hat{s}_t\right)$ requires backpropagating gradients through the policy network, which is both expensive to compute and can lead to high variance on noisy, out-of-distribution states.
Due to this, we apply policy guidance to only the noised action, yielding our policy-guided score function
\begin{align}
    s_{\mkern 1mu \text{PGD}}(\bm{\hat{\tau}};\sigma) = \underbrace{\nabla_{\bm{\hat{\tau}}}\log \poff(\bm{\hat{\tau}};\sigma)}_{\mathclap{\text{Behavior score function}}} + \underbrace{\nabla_{\bm{\hat{a}}} \log \qtarg(\bm{\hat{\tau}})}_{\mathclap{\text{Target policy guidance}}} \text{,} \label{eq:pgd-score}
\end{align}
where (abusing notation) $\nabla_{\bm{\hat{a}}}$ denotes the gradient $\nabla_{\bm{\hat{\tau}}}$ of $\bm{\hat{\tau}}$, with non-action components set to 0.

\subsection{\label{sec:stable-guidance}Improving Policy Guidance}

\begin{figure}[t]
\centering
\vspace{-2mm}
\includegraphics[width=\textwidth]{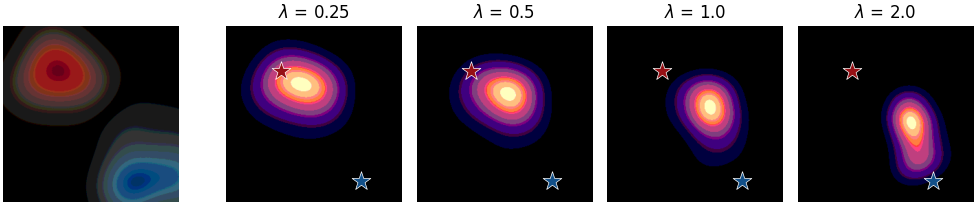}
\vspace{-2mm}
\caption{
\textbf{Left:} Trajectory probability distribution for an example \textcolor{red}{behavior distribution $\poff(\bm{\tau})$} and \textcolor{blue}{target policy likelihood $\qtarg(\bm{\tau})$}.
\textbf{Right:} Corresponding PGD sampling distribution (\autoref{eq:pgd-dist}) computed over a range of policy-guidance coefficients $\lambda$.
By increasing $\lambda$, we transform from the sampling distribution towards the regions of high target policy likelihood, making PGD an effective mechanism for controlling the level of regularization towards the behavior distribution.
}
\label{fig:coeff-vis}
\end{figure}

\paragraph{Controlling Guidance Strength}
A standard technique from classifier-guided diffusion is the use of guidance coefficients~\citep{dhariwal2021diffusion}.
These augment the guided score function by introducing a controllable coefficient on the guidance term. Applied to the PGD score function (\autoref{eq:pgd-score}), this has the form
\begin{equation}
    s_{\mkern 1mu \text{PGD}}(\bm{\hat{\tau}};\sigma, \lambda) = \nabla_{\bm{\hat{\tau}}}\log \poff(\bm{\hat{\tau}};\sigma) + \lambda \nabla_{\bm{\hat{a}}} \log \qtarg(\bm{\hat{\tau}}) \text{,}
\end{equation}
where $\lambda$ denotes the guidance coefficient. As $\sigma \to 0$, this transforms the sampling distribution to
\begin{equation}
    \mathcal{F}(\bm{\tau} | \pi_{\text{target}};\lambda) \propto \poff(\bm{\tau}) \cdot \qtarg(\bm{\tau})^{\lambda} \text{.}
    \label{eq:pgd-dist}
\end{equation}
Intuitively, $\lambda$ interpolates the actions in the sampling distribution between the behavior and target distributions.
By tuning $\lambda$, we can therefore control the strength of guidance towards the target policy, avoiding high dynamics error when the target policy is far from the behavior policy.
We visualize this effect in \autoref{fig:coeff-vis} and analyze its impact on target policy likelihood in \autoref{fig:action-prob}.

Following \citet{ma2023elucidating}, we also apply a cosine guidance schedule to the guidance coefficient,
\begin{equation}
    \lambda_n = \lambda \cdot (\sigma_n + \beta \sigma_N \cdot \sin(\pi \cdot n/N))\text{,}
\end{equation}
where $\beta$ is the cosine weight, which is set to 0.3 in all experiments.
By decreasing the strength of guidance in later steps, we find that this schedule stabilizes guidance and reduces dynamics error.

\paragraph{Stabilizing Guided Diffusion}
When under distribution shift, RL policies are known to suffer from poor generalization to unseen states~\citep{Kirk_2023}. This makes policy guidance challenging, since the policy must operate on noised states, and compute action gradients from noised actions. Similar issues have been studied in classifier-guided diffusion~\citep{ma2023elucidating}, where the classifier gradient can be unstable when exposed to out-of-distribution inputs. \citet{bansal2023universal} alleviate this issue by applying guidance to the denoised sample estimated by the denoiser model, rather than the original noised sample, in addition to normalizing the guidance gradient to a unit vector. By applying these techniques to policy guidance, we lessen the need for the target policy to generalize to noisy states, which we find decreases dynamics error.

\newcommand{\none}{\color{black}---}
\newcommand{\low}{\color{ForestGreen}Low}
\newcommand{\high}{\color{red}High}
\newcommand{\pmax}{\color{black}---}
\newcommand{\plow}{\color{red}Low}
\newcommand{\phigh}{\color{ForestGreen}High}
\renewcommand{\arraystretch}{1.1}
\begin{table}[hbt]
    \centering
    \caption{Overview of training experience sources in offline RL---for each, we consider the sampling distribution, expected \textbf{error} in transition dynamics, \textbf{likelihood} of actions under the target policy, and state space \textbf{coverage} beyond the behavior distribution. Policy-guided diffusion provides an effective trade-off between each error, likelihood, and coverage.}
    \begin{tabular}{@{}llccc@{}}
        \toprule
            \textbf{Data source} & \textbf{Distribution} & \textbf{Error ($\bm\downarrow$)} & \textbf{Likelihood ($\bm\uparrow$)} & \textbf{Coverage ($\bm\uparrow$)}\\
        \midrule
        Offline dataset & $\poff(\bm{\tau})$ & \none{} & \plow{} & \plow{} \\
        \cmidrule{2-5}
        Episodic world model & $\ptarg(\bm{\tau})$ & \high{} & \pmax{} & \phigh{} \\
        Truncated world model & \autoref{eq:trunc-wm} & \low{} & \pmax{} & \plow{} \\
        \cmidrule{2-5}
        Unguided diffusion & $\poff(\bm{\tau})$ & \low{} & \plow{} & \phigh{} \\
        Policy-guided diffusion & \autoref{eq:pgd} & \low{} & \phigh{} & \phigh{} \\
        \bottomrule
    \end{tabular}
    \label{tab:methods}
\end{table}
\renewcommand{\arraystretch}{1}

\section{Results\label{sec:results}}
Through our experiments, we first demonstrate that agents trained with synthetic experience from PGD outperform those trained on unguided synthetic data or directly on the offline dataset (\autoref{sec:rl-results}). We show that this effect is consistent across agents (TD3+BC and IQL), environments (HalfCheetah, Walker2d, Hopper, and Maze), behavior policies (random, mixed, and medium), and modes of data generation (continuous and periodic).
Following this, we demonstrate that tuning the guidance coefficient enables PGD to sample trajectories with high action likelihood across a range of target policies.
Finally, we verify that PGD retains low dynamics error despite sampling high-likelihood actions from the policy (\autoref{sec:analysis}).

\subsection{Experimental Setup}
We evaluate PGD on the MuJoCo and Maze2d continuous control datasets from D4RL~\citep{fu2020d4rl, mujoco}.
For MuJoCo, we consider the HalfCheetah, Walker2d, and Hopper environments with random (randomly initialized behavior policy), medium (suboptimal behavior policy), and medium-replay (or ``mixed'', the replay buffer from medium policy training) datasets. 
For Maze2d we consider the original (sparse reward) instances of the umaze, medium and large layouts.
We train 4 trajectory diffusion models on each dataset, for which we detail hyperparameters in \autoref{sec:all-hypers}.
In \autoref{sec:analysis}, we conduct analysis of PGD against MOPO-style PETS~\citep{chua2018deep} models, an autoregressive world model composed of an ensemble of probabilistic models, for which we use model weights from OfflineRL-Kit~\citep{offinerlkit}.

To demonstrate synthetic experience from PGD as a drop-in substitute for the real dataset, we transfer the original hyperparameters for IQL~\citep{kostrikov2021offline} and TD3+BC~\citep{fujimoto2021minimalist}---as tuned on the real datasets---\textit{without any further tuning}.
Policy guidance requires a stochastic target policy, in order to compute the gradient of the action distribution.
Since TD3+BC trains a deterministic policy, we perform guidance by modeling the action distribution as a unit Gaussian centered on the deterministic action.
We implement all agents and diffusion models from scratch in Jax~\citep{jax2018github}, which may be found at \href{https://github.com/EmptyJackson/policy-guided-diffusion}{https://github.com/EmptyJackson/policy-guided-diffusion}.

\subsection{Offline Reinforcement Learning\label{sec:rl-results}}
For each D4RL dataset, we train two popular model-free offline algorithms, TD3+BC~\citep{fujimoto2021minimalist} and IQL~\citep{kostrikov2021offline} on synthetic experience generated by trajectory diffusion models with and without policy guidance, as well as on the real dataset.
We first consider \textit{periodic generation} of synthetic data, in which the synthetic dataset is regenerated after extended periods of agent training, such that the agent is near convergence on the synthetic dataset at the point it is regenerated with the current policy.
Each epoch, we generate a dataset of $2^{14}$ synthetic trajectories of length 16.
Following the notation of \autoref{alg:training}, we set the number of epochs to $N_\text{epochs} = 4$ with $N_\text{policy} = \num[group-separator={,}]{250000}$ train steps per epoch, meaning the agent is trained to close to convergence before the dataset is regenerated.
This can be viewed as solving a sequence of offline RL tasks with synthetic datasets, in which the behavior policy is the target policy from the previous generation.

Using periodic generation, performance improves significantly across benchmarks for both IQL and TD3+BC (\autoref{tab:delayed-results}).
In MuJoCo, the most consistent improvement is on mixed datasets, with 4 out of 6 experiments achieving significant performance improvement.
This is to be expected, as these datasets contain experience from a mixture of behavior policy levels.
In this case, the diffusion model is likely to be able to represent a wide variety of policies, and on-policy guidance would naturally produce higher return trajectories as the target policy improves.

\setlength{\tabcolsep}{4pt}
\renewcommand{\arraystretch}{1.}
\begin{table}[htb]
    \small
    \centering
    \caption{Final return of IQL and TD3+BC agents trained on real, unguided ($\lambda = 0$) synthetic and policy-guided ($\lambda = 1$) synthetic data---mean and standard error over 4 seeds (diffusion models and agents) is presented, with significant improvements ($p < 0.05$) shaded.}
    \begin{tabular}{@{}llc@{\hspace{.05mm}}NNNc@{\hspace{.05mm}}|c@{\hspace{.05mm}}NNN@{}}
        \toprule
        \multicolumn{2}{c}{} & \multicolumn{11}{c}{\textbf{IQL}} && \multicolumn{9}{c}{\textbf{TD3+BC}} \\
        \cmidrule{4-12} \cmidrule{15-23}
        \multicolumn{2}{c}{} && \multicolumn{3}{c}{\textbf{Dataset}} & \multicolumn{3}{c}{\hspace{2mm}\textbf{Unguided}} & \multicolumn{3}{c}{\hspace{2mm}\textbf{Guided}} &\multicolumn{2}{c}{}& \multicolumn{3}{c}{\textbf{Dataset}} & \multicolumn{3}{c}{\textbf{Unguided}} & \multicolumn{3}{c}{\hspace{2mm}\textbf{Guided}} \\
        \midrule
        \multirow{3}{*}{\STAB{\rotatebox[origin=c]{90}{Random}}} & HalfCheetah &&
            $\mhl{9.1} $&$\pm$&$ 2.2$ & $2.6 $&$\pm$&$ 0.1$ & $\mhl{6.5} $&$\pm$&$ 1.7$ &&&
            $11.2 $&$\pm$&$ 0.8$ & $11.0 $&$\pm$&$ 0.4$ & $\mhl{21.1} $&$\pm$&$ 0.9$ \\
        & Walker2d &&
            $4.3 $&$\pm$&$ 0.5$ & $2.7 $&$\pm$&$ 0.7$ & $\mhl{5.3} $&$\pm$&$ 0.3$ &&&
            $0.5 $&$\pm$&$ 0.3$ & $1.1 $&$\pm$&$ 1.2$ & $-0.3 $&$\pm$&$ 0.1$ \\
        & Hopper &&
            $\mhl{7.4} $&$\pm$&$ 0.4$ & $5.2 $&$\pm$&$ 0.9$ & $4.9 $&$\pm$&$ 1.0$ &&&
            $7.4 $&$\pm$&$ 0.6$ & $4.2 $&$\pm$&$ 1.4$ & $5.5 $&$\pm$&$ 2.1$ \\
        \cmidrule{2-23}
        \multirow{3}{*}{\STAB{\rotatebox[origin=c]{90}{Mixed}}} & HalfCheetah &&
            $\mhl{44.2} $&$\pm$&$ 0.2$ & $43.6 $&$\pm$&$ 0.2$ & $43.6 $&$\pm$&$ 0.2$ &&&
            $44.7 $&$\pm$&$ 0.1$ & $43.1 $&$\pm$&$ 0.2$ & $\mhl{46.1} $&$\pm$&$ 0.3$ \\
        & Walker2d &&
            $81.3 $&$\pm$&$ 2.0$ & $\mhl{85.2} $&$\pm$&$ 0.3$ & $\mhl{84.9} $&$\pm$&$ 1.4$ &&&
            $82.7 $&$\pm$&$ 1.3$ & $70.7 $&$\pm$&$ 10.1$ & $84.0 $&$\pm$&$ 1.0$ \\
        & Hopper &&
            $82.9 $&$\pm$&$ 3.5$ & $\mhl{97.4} $&$\pm$&$ 2.7$ & $\mhl{100.5} $&$\pm$&$ 0.5$ &&&
            $58.6 $&$\pm$&$ 11.2$ & $52.1 $&$\pm$&$ 1.8$ & $\mhl{91.9} $&$\pm$&$ 4.3$ \\
        \cmidrule{2-23}
        \multirow{3}{*}{\STAB{\rotatebox[origin=c]{90}{Medium}}} & HalfCheetah &&
            $\mhl{48.4} $&$\pm$&$ 0.1$ & $45.4 $&$\pm$&$ 0.1$ & $45.1 $&$\pm$&$ 0.1$ &&&
            $\mhl{48.6} $&$\pm$&$ 0.1$ & $45.3 $&$\pm$&$ 0.2$ & $47.6 $&$\pm$&$ 0.3$ \\
        & Walker2d &&
            $81.7 $&$\pm$&$ 1.4$ & $82.1 $&$\pm$&$ 0.9$ & $77.8 $&$\pm$&$ 3.6$ &&&
            $84.8 $&$\pm$&$ 0.1$ & $85.2 $&$\pm$&$ 0.2$ & $\mhl{86.3} $&$\pm$&$ 0.3$ \\
        & Hopper &&
            $63.6 $&$\pm$&$ 0.8$ & $59.7 $&$\pm$&$ 2.0$ & $62.8 $&$\pm$&$ 1.2$ &&&
            $\mhl{62.4} $&$\pm$&$ 0.9$ & $57.4 $&$\pm$&$ 0.4$ & $\mhl{63.1} $&$\pm$&$ 0.6$ \\
        \cmidrule{2-23}
         & Total &&
            $46.9 $&$\pm$&$ 0.4$ & $47.0 $&$\pm$&$ 0.4$ & $\mhl{47.9} $&$\pm$&$ 0.3$ &&&
            $44.5 $&$\pm$&$ 1.1$ & $41.1 $&$\pm$&$ 1.1$ & $\mhl{49.5} $&$\pm$&$ 0.9$ \\
        \midrule
        \multirow{3}{*}{\STAB{\rotatebox[origin=c]{90}{Maze2d}}} & UMaze &&
            $42.6 $&$\pm$&$ 0.4$ & $42.9 $&$\pm$&$ 1.8$ & $43.8 $&$\pm$&$ 3.5$ &&&
            $50.0 $&$\pm$&$ 2.4$ & $33.8 $&$\pm$&$ 3.0$ & $\mhl{76.2} $&$\pm$&$ 17.4$ \\
        & Medium &&
            $38.5 $&$\pm$&$ 1.9$ & $33.4 $&$\pm$&$ 3.2$ & $\mhl{60.0} $&$\pm$&$ 13.9$ &&&
            $32.1 $&$\pm$&$ 6.8$ & $24.0 $&$\pm$&$ 4.0$ & $\mhl{89.6} $&$\pm$&$ 19.9$ \\
        & Large &&
            $\mhl{50.9} $&$\pm$&$ 5.8$ & $23.4 $&$\pm$&$ 8.0$ & $\mhl{45.3} $&$\pm$&$ 14.8$ &&&
            $137.2 $&$\pm$&$ 20.2$ & $93.3 $&$\pm$&$ 31.0$ & $131.1 $&$\pm$&$ 37.5$ \\
        \cmidrule{2-23}
         & Total &&
            $\mhl{44.0} $&$\pm$&$ 2.2$ & $33.2 $&$\pm$&$ 1.8$ & $\mhl{49.7} $&$\pm$&$ 9.5$ &&&
            $73.1 $&$\pm$&$ 6.7$ & $50.4 $&$\pm$&$ 11.1$ & $\mhl{99.0} $&$\pm$&$ 14.5$ \\
        \bottomrule
    \end{tabular}
    \label{tab:delayed-results}
\end{table}
\renewcommand{\arraystretch}{1}

In order to demonstrate the flexibility of PGD, we also evaluate PGD in a \textit{continuous generation} setting, using a data generation rate closer to that of traditional model-based methods. For this, we set $N_\text{epochs} = \num[group-separator={,}]{100}$ and $N_\text{policy} = \num[group-separator={,}]{10000}$, then lower the sample size to match the overall number of synthetic trajectories generated by periodic generation across training. Due to the decrease in sample size, we maintain each generated dataset across epochs in a replay buffer, with each dataset being removed after 10 epochs.

\begin{figure}[h]
    \centering
    \begin{subfigure}{\linewidth}
        \centering
        \includegraphics[width=\linewidth]{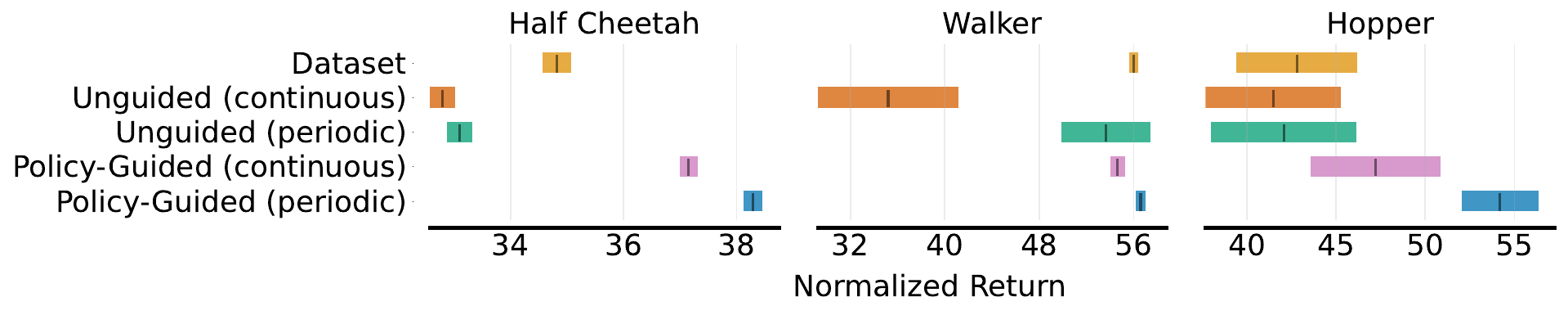}
        \vspace{-1.75em}
        \caption{Environments}
    \end{subfigure}
    \vspace{.75em}
    \newline
    \begin{subfigure}{\linewidth}
        \centering
        \includegraphics[width=\linewidth]{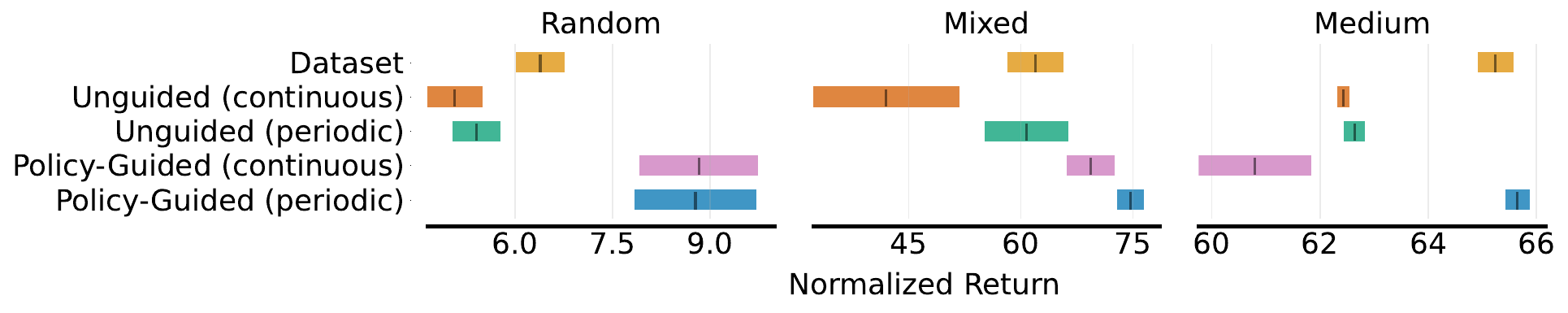}
        \vspace{-1.75em}
        \caption{Behavior policies}
    \end{subfigure}
    \caption{Aggregate MuJoCo performance after training on unguided or policy-guided synthetic data under continuous and periodic dataset generation, as well as on the real dataset. For each setting, mean return over TD3+BC and IQL agents is marked, with standard error over 4 seeds (diffusion models and agents) highlighted.}
    \label{fig:agg-scores}
\end{figure}

We see similar improvements in performance against real and unguided synthetic data under this approach, with PGD outperforming real data on 2 out of 3 environments and datasets (\autoref{fig:agg-scores}).
Periodic generation outperforms continuous generation across environments and behavior policies, which we attribute to training stability, especially when performing guidance early in training.
Regardless, both approaches consistently outperform training on real and unguided synthetic data, demonstrating the potential of PGD as a drop-in extension to replay and model-based RL methods.

\subsection{\label{sec:analysis}Synthetic Trajectory Analysis}

We now analyze the quality of trajectories produced by PGD against those from unguided diffusion and autoregressive world model (PETS) rollouts.
In principle, we seek to evaluate the divergence of these sampling distributions from the true target distribution.
However, this is not tractable to compute directly, so we instead investigate two proxy objectives:
\begin{enumerate}
    \item \textbf{Trajectory Likelihood}: mean log-likelihood of actions under the target policy; and
    \item \textbf{Dynamics Error}: mean squared error between states in the synthetic trajectory and real environment, when rolled out with the same initial state and action sequence.
\end{enumerate}

In our experiments, we consider trajectory diffusion and MOPO-style PETS~\citep{chua2018deep} models trained on representative datasets from the D4RL~\citep{fu2020d4rl} benchmark that were featured in the previous section.
Specifically, we consider the models trained on halfcheetah-medium, before sampling trajectories with IQL target policies trained on the halfcheetah-random, -medium, and -expert.
This enables us to test the robustness of these models to target policies far from the behavior policy, both in performance and policy entropy.

\begin{figure}[ht]
    \centering
    \includegraphics[width=\linewidth]{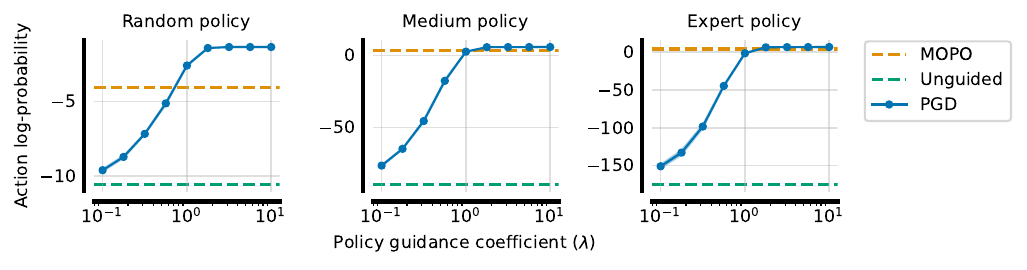}
    \caption{Action probability of synthetic trajectories generated by diffusion and PETS models trained on halfcheetah-medium. Target policies are trained on halfcheetah-random, halfcheetah-medium, and halfcheetah-expert datasets, demonstrating robustness to OOD actions. Standard error over 4 diffusion model seeds is shaded (but negligible), with mean computed over 2048 synthetic trajectories.}
    \label{fig:action-prob}
\end{figure}

\paragraph{Policy Guidance Increases Trajectory Likelihood} 
In \autoref{fig:action-prob}, we present the trajectory likelihood of synthetic trajectories over varying degrees of guidance.
Unsurprisingly, unguided diffusion generates low probability trajectories for all target policies, due to it directly modeling the behavior distribution.
However, as we increase the guidance coefficient $\lambda$, trajectory likelihood increases monotonically under each target policy.
Furthermore, this effect is robust across target policies, giving the ability to sample high-probability trajectories with OOD target policies.
The value of $\lambda$ required to achieve the same action likelihood as direct action sampling (PETS) varies with the target policy.
Since this threshold increases with target policy performance, we hypothesize that it increases with target policy entropy.
Based on this, a promising avenue for future work is automatically tuning $\lambda$ for hyperparameter-free guidance.

\begin{figure}[t]
    \centering
    \includegraphics[width=\linewidth]{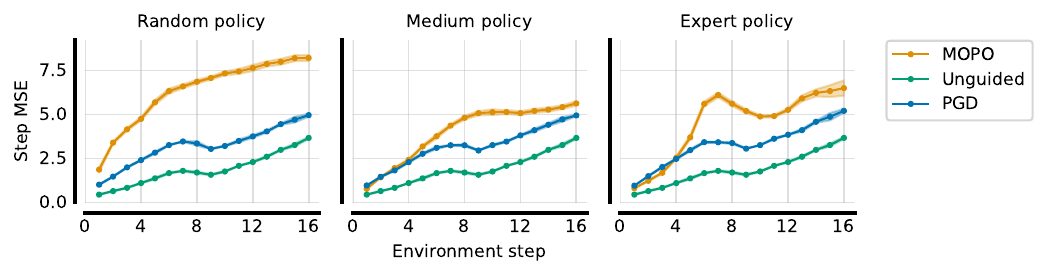}
    \caption{Dynamics mean squared error of synthetic trajectories generated by diffusion and PETS models trained on halfcheetah-medium.
    Standard error over 4 diffusion model seeds and 3 PETS seeds (via OfflineRL-Kit) is shaded, with each generating 2048 synthetic trajectories for analysis.
    }
    \label{fig:traj-mse}
\end{figure}

\paragraph{Policy Guided Diffusion Achieves Lower Error Than Autoregressive Models}
In \autoref{fig:traj-mse}, we present the dynamics error of synthetic trajectories over 16 rollout steps.
For a fair comparison, we fix the guidance coefficient of PGD to $\lambda = 1.0$, since this was sufficient to match the trajectory likelihood of PETS (\autoref{fig:action-prob}).
Over all target policies, PGD achieves significantly lower error than PETS.
Furthermore, PGD has similar levels of error across target policies, while PETS suffers from significantly higher error on OOD (random and expert) target policies.
This highlights the robustness of PGD to target policy, a critical feature for generating high-likelihood training data throughout tabula rasa policy training.

\section{Related Work}
\label{sec:related_work}

\paragraph{Model-based Offline Reinforcement Learning}
Model-based methods in offline RL~\citep{mopo, kidambi2020morel, rambo, lu2022revisiting} are designed to augment the offline buffer with additional on-policy samples in order to mitigate distribution shift.
This is typically done by rolling out a policy in a learned world model~\citep{janner2019mbpo} and applying a suitable pessimism term in order to account for dynamics model errors.
While these methods share the same overall motivation as our paper, the empirical realization is quite different.
In particular, forward dynamics models are liable to compounding errors over long horizons, resulting in model exploitation, whereas our trajectories are generated in a single step.

\paragraph{Model-free Offline Reinforcement Learning}
Model-free methods in offline RL typically tackle the out-of-sample issue by applying conservatism to the value function or by constraining the policy to remain close to the data.
For example, CQL~\citep{kumar2020conservative} and EDAC~\citep{an2021uncertainty} both aim to minimize the values of out-of-distribution actions.
Meanwhile, BCQ~\citep{bcq} ensures that actions used in value targets are in-distribution with the behavioral policy using constrained optimization.
We take the opposite approach in this paper: by enabling our diffusion model to generate on-policy samples without diverging from the behavior distribution, we reduce the need for conservatism.

\paragraph{Diffusion in Reinforcement Learning}
Diffusion models are a flexible method for data augmentation in reinforcement learning.
SynthER~\citep{lu2023synthetic} uses unguided diffusion models to upsample offline or online RL datasets, which are then used by model-free off-policy algorithms.
While this improves performance, SynthER uses unguided diffusion to model the behavior distribution, resulting in the same issue of distributional shift.
Similarly, MTDiff~\citep{he2023diffusion} considers unguided data generation in multitask settings.

Diffusion models have also been used to train world models. \citet{zhang2023learning} train a world model for sensor observations by first tokenizing using VQ-VAE and then predicting future observations via discrete diffusion. \citet{alonso2023diffusion} also train a world model using diffusion and demonstrate it can more accurately predict future observations. However, neither of these approaches model the whole trajectory, thereby suffering from compounding error, nor do they apply policy guidance.
Parallel to this work, \citet{rigter2023world} use guidance from a policy to augment a diffusion world model for online RL. By contrast, we focus on the offline RL setting, provide a theoretical derivation and motivation for the trajectory distribution modeled by policy guidance, and demonstrate improvements in downstream policy performance.

Diffusion models are also used elsewhere in reinforcement learning.
For example, Diffuser~\citep{janner2022diffuser} and Decision Diffuser~\citep{ajay2023is} use trajectory diffusion models for planning and to bias planned trajectories towards high return.
By contrast, we use on-policy guidance and train on the generated data.
Diffusion models have also been used as an expressive policy class~\citep{wang2023diffusion} for $Q$-learning, showing improvement over MLPs.

\section{Conclusion}
\label{conclusion}
We presented policy-guided diffusion, a method for controllable generation of synthetic trajectories in offline RL.
We provided a theoretical analysis of existing approaches to synthetic experience generation, identifying the advantages of direct trajectory generation compared to autoregressive methods.
Motivated by this, we proposed PGD under the direct approach, deriving the regularized target distribution modeled by policy guidance.

Evaluating against PETS deep ensembles, a state-of-the-art autoregressive approach, we found that PGD can generate synthetic experience at the same target policy likelihood with significantly lower dynamics error.
Furthermore, we found consistent improvements in downstream agent performance over a range of environments and behavior policies when trained on policy-guided synthetic data, against real and unguided synthetic experience.

By addressing the out-of-sample issue through synthetic data, we hope that this work enables the development of less conservative algorithms for offline RL.
There are a range of promising avenues for future work, including automatically tuning the guidance coefficient for hyperparameter-free guidance, leveraging on-policy RL techniques with policy-guided data, and extending this approach to large-scale video generation models.


\subsubsection*{Acknowledgments}
\label{sec:ack}
We thank Mattie Fellows and Sebastian Towers for their insights regarding our method's theoretical underpinning, as well as Alex Goldie and the NeurIPS 2023 Robot Learning Workshop reviewers for their helpful feedback.
Matthew Jackson is funded by the EPSRC Centre for Doctoral Training in Autonomous Intelligent Machines and Systems, and Amazon Web Services.


\bibliography{main}

\begin{thebibliography}{38}
\providecommand{\natexlab}[1]{#1}
\providecommand{\url}[1]{\texttt{#1}}
\expandafter\ifx\csname urlstyle\endcsname\relax
  \providecommand{\doi}[1]{doi: #1}\else
  \providecommand{\doi}{doi: \begingroup \urlstyle{rm}\Url}\fi

\bibitem[Ajay et~al.(2023)Ajay, Du, Gupta, Tenenbaum, Jaakkola, and Agrawal]{ajay2023is}
Anurag Ajay, Yilun Du, Abhi Gupta, Joshua~B. Tenenbaum, Tommi~S. Jaakkola, and Pulkit Agrawal.
\newblock Is conditional generative modeling all you need for decision making?
\newblock In \emph{The Eleventh International Conference on Learning Representations}, 2023.
\newblock URL \url{https://openreview.net/forum?id=sP1fo2K9DFG}.

\bibitem[Alonso et~al.(2023)Alonso, Jelley, Kanervisto, and Pearce]{alonso2023diffusion}
Eloi Alonso, Adam Jelley, Anssi Kanervisto, and Tim Pearce.
\newblock Diffusion world models.
\newblock 2023.

\bibitem[An et~al.(2021)An, Moon, Kim, and Song]{an2021uncertainty}
Gaon An, Seungyong Moon, Jang-Hyun Kim, and Hyun~Oh Song.
\newblock Uncertainty-based offline reinforcement learning with diversified q-ensemble.
\newblock \emph{Advances in neural information processing systems}, 34:\penalty0 7436--7447, 2021.

\bibitem[Ball et~al.(2021)Ball, Lu, Parker-Holder, and Roberts]{augwm}
Philip~J Ball, Cong Lu, Jack Parker-Holder, and Stephen Roberts.
\newblock Augmented world models facilitate zero-shot dynamics generalization from a single offline environment.
\newblock In Marina Meila and Tong Zhang (eds.), \emph{Proceedings of the 38th International Conference on Machine Learning}, volume 139 of \emph{Proceedings of Machine Learning Research}, pp.\  619--629. PMLR, 18--24 Jul 2021.

\bibitem[Bansal et~al.(2023)Bansal, Chu, Schwarzschild, Sengupta, Goldblum, Geiping, and Goldstein]{bansal2023universal}
Arpit Bansal, Hong-Min Chu, Avi Schwarzschild, Soumyadip Sengupta, Micah Goldblum, Jonas Geiping, and Tom Goldstein.
\newblock Universal guidance for diffusion models.
\newblock In \emph{Proceedings of the IEEE/CVF Conference on Computer Vision and Pattern Recognition}, pp.\  843--852, 2023.

\bibitem[Bradbury et~al.(2018)Bradbury, Frostig, Hawkins, Johnson, Leary, Maclaurin, Necula, Paszke, Vander{P}las, Wanderman-{M}ilne, and Zhang]{jax2018github}
James Bradbury, Roy Frostig, Peter Hawkins, Matthew~James Johnson, Chris Leary, Dougal Maclaurin, George Necula, Adam Paszke, Jake Vander{P}las, Skye Wanderman-{M}ilne, and Qiao Zhang.
\newblock {JAX}: composable transformations of {P}ython+{N}um{P}y programs, 2018.
\newblock URL \url{http://github.com/google/jax}.

\bibitem[Chua et~al.(2018)Chua, Calandra, McAllister, and Levine]{chua2018deep}
Kurtland Chua, Roberto Calandra, Rowan McAllister, and Sergey Levine.
\newblock Deep reinforcement learning in a handful of trials using probabilistic dynamics models, 2018.

\bibitem[Dhariwal \& Nichol(2021)Dhariwal and Nichol]{dhariwal2021diffusion}
Prafulla Dhariwal and Alexander Nichol.
\newblock Diffusion models beat gans on image synthesis.
\newblock \emph{Advances in neural information processing systems}, 34:\penalty0 8780--8794, 2021.

\bibitem[Fu et~al.(2020)Fu, Kumar, Nachum, Tucker, and Levine]{fu2020d4rl}
Justin Fu, Aviral Kumar, Ofir Nachum, George Tucker, and Sergey Levine.
\newblock D4rl: Datasets for deep data-driven reinforcement learning.
\newblock \emph{arXiv preprint arXiv:2004.07219}, 2020.

\bibitem[Fujimoto \& Gu(2021)Fujimoto and Gu]{fujimoto2021minimalist}
Scott Fujimoto and Shixiang~Shane Gu.
\newblock A minimalist approach to offline reinforcement learning.
\newblock In \emph{Thirty-Fifth Conference on Neural Information Processing Systems}, 2021.

\bibitem[Fujimoto et~al.(2019)Fujimoto, Meger, and Precup]{bcq}
Scott Fujimoto, David Meger, and Doina Precup.
\newblock Off-policy deep reinforcement learning without exploration.
\newblock In Kamalika Chaudhuri and Ruslan Salakhutdinov (eds.), \emph{Proceedings of the 36th International Conference on Machine Learning}, volume~97 of \emph{Proceedings of Machine Learning Research}, pp.\  2052--2062. PMLR, 09--15 Jun 2019.
\newblock URL \url{https://proceedings.mlr.press/v97/fujimoto19a.html}.

\bibitem[He et~al.(2023)He, Bai, Xu, Yang, Zhang, Wang, Zhao, and Li]{he2023diffusion}
Haoran He, Chenjia Bai, Kang Xu, Zhuoran Yang, Weinan Zhang, Dong Wang, Bin Zhao, and Xuelong Li.
\newblock Diffusion model is an effective planner and data synthesizer for multi-task reinforcement learning.
\newblock \emph{arXiv preprint arXiv:2305.18459}, 2023.

\bibitem[Ho et~al.(2020)Ho, Jain, and Abbeel]{ddpm}
Jonathan Ho, Ajay Jain, and Pieter Abbeel.
\newblock Denoising diffusion probabilistic models.
\newblock In H.~Larochelle, M.~Ranzato, R.~Hadsell, M.F. Balcan, and H.~Lin (eds.), \emph{Advances in Neural Information Processing Systems}, volume~33, pp.\  6840--6851. Curran Associates, Inc., 2020.
\newblock URL \url{https://proceedings.neurips.cc/paper/2020/file/4c5bcfec8584af0d967f1ab10179ca4b-Paper.pdf}.

\bibitem[Hyv{\"a}rinen \& Dayan(2005)Hyv{\"a}rinen and Dayan]{hyvarinen2005estimation}
Aapo Hyv{\"a}rinen and Peter Dayan.
\newblock Estimation of non-normalized statistical models by score matching.
\newblock \emph{Journal of Machine Learning Research}, 6\penalty0 (4), 2005.

\bibitem[Janner et~al.(2019)Janner, Fu, Zhang, and Levine]{janner2019mbpo}
Michael Janner, Justin Fu, Marvin Zhang, and Sergey Levine.
\newblock When to trust your model: Model-based policy optimization.
\newblock In \emph{Advances in Neural Information Processing Systems}, 2019.

\bibitem[Janner et~al.(2022)Janner, Du, Tenenbaum, and Levine]{janner2022diffuser}
Michael Janner, Yilun Du, Joshua Tenenbaum, and Sergey Levine.
\newblock Planning with diffusion for flexible behavior synthesis.
\newblock In \emph{International Conference on Machine Learning}, 2022.

\bibitem[Karras et~al.(2022)Karras, Aittala, Aila, and Laine]{karras2022elucidating}
Tero Karras, Miika Aittala, Timo Aila, and Samuli Laine.
\newblock Elucidating the design space of diffusion-based generative models.
\newblock \emph{Advances in Neural Information Processing Systems}, 35:\penalty0 26565--26577, 2022.

\bibitem[Kidambi et~al.(2020)Kidambi, Rajeswaran, Netrapalli, and Joachims]{kidambi2020morel}
Rahul Kidambi, Aravind Rajeswaran, Praneeth Netrapalli, and Thorsten Joachims.
\newblock Morel: Model-based offline reinforcement learning.
\newblock \emph{Advances in neural information processing systems}, 33:\penalty0 21810--21823, 2020.

\bibitem[Kirk et~al.(2023)Kirk, Zhang, Grefenstette, and Rocktäschel]{Kirk_2023}
Robert Kirk, Amy Zhang, Edward Grefenstette, and Tim Rocktäschel.
\newblock A survey of zero-shot generalisation in deep reinforcement learning.
\newblock \emph{Journal of Artificial Intelligence Research}, 76:\penalty0 201–264, January 2023.
\newblock ISSN 1076-9757.
\newblock \doi{10.1613/jair.1.14174}.
\newblock URL \url{http://dx.doi.org/10.1613/jair.1.14174}.

\bibitem[Kostrikov et~al.(2021)Kostrikov, Nair, and Levine]{kostrikov2021offline}
Ilya Kostrikov, Ashvin Nair, and Sergey Levine.
\newblock Offline reinforcement learning with implicit q-learning.
\newblock \emph{arXiv preprint arXiv:2110.06169}, 2021.

\bibitem[Kumar et~al.(2020)Kumar, Zhou, Tucker, and Levine]{kumar2020conservative}
Aviral Kumar, Aurick Zhou, George Tucker, and Sergey Levine.
\newblock Conservative q-learning for offline reinforcement learning.
\newblock \emph{Advances in Neural Information Processing Systems}, 33:\penalty0 1179--1191, 2020.

\bibitem[Levine et~al.(2020)Levine, Kumar, Tucker, and Fu]{levine2020offline}
Sergey Levine, Aviral Kumar, George Tucker, and Justin Fu.
\newblock Offline reinforcement learning: Tutorial, review, and perspectives on open problems.
\newblock \emph{arXiv preprint arXiv:2005.01643}, 2020.

\bibitem[Lu et~al.(2022)Lu, Ball, Parker-Holder, Osborne, and Roberts]{lu2022revisiting}
Cong Lu, Philip Ball, Jack Parker-Holder, Michael Osborne, and Stephen~J. Roberts.
\newblock Revisiting design choices in offline model based reinforcement learning.
\newblock In \emph{International Conference on Learning Representations}, 2022.
\newblock URL \url{https://openreview.net/forum?id=zz9hXVhf40}.

\bibitem[Lu et~al.(2023)Lu, Ball, Teh, and Parker-Holder]{lu2023synthetic}
Cong Lu, Philip~J. Ball, Yee~Whye Teh, and Jack Parker-Holder.
\newblock Synthetic experience replay.
\newblock In \emph{Thirty-seventh Conference on Neural Information Processing Systems}, 2023.
\newblock URL \url{https://openreview.net/forum?id=6jNQ1AY1Uf}.

\bibitem[Ma et~al.(2023)Ma, Hu, Wang, and Sun]{ma2023elucidating}
Jiajun Ma, Tianyang Hu, Wenjia Wang, and Jiacheng Sun.
\newblock Elucidating the design space of classifier-guided diffusion generation.
\newblock \emph{arXiv preprint arXiv:2310.11311}, 2023.

\bibitem[Precup(2000)]{precup2000eligibility}
Doina Precup.
\newblock Eligibility traces for off-policy policy evaluation.
\newblock \emph{Computer Science Department Faculty Publication Series}, pp.\ ~80, 2000.

\bibitem[Precup et~al.(2000)Precup, Sutton, and Singh]{10.5555/645529.658134}
Doina Precup, Richard~S. Sutton, and Satinder~P. Singh.
\newblock Eligibility traces for off-policy policy evaluation.
\newblock In \emph{Proceedings of the Seventeenth International Conference on Machine Learning}, ICML '00, pp.\  759–766, San Francisco, CA, USA, 2000. Morgan Kaufmann Publishers Inc.
\newblock ISBN 1558607072.

\bibitem[Rigter et~al.(2022)Rigter, Lacerda, and Hawes]{rambo}
Marc Rigter, Bruno Lacerda, and Nick Hawes.
\newblock {RAMBO}-{RL}: Robust adversarial model-based offline reinforcement learning.
\newblock In Alice~H. Oh, Alekh Agarwal, Danielle Belgrave, and Kyunghyun Cho (eds.), \emph{Advances in Neural Information Processing Systems}, 2022.
\newblock URL \url{https://openreview.net/forum?id=nrksGSRT7kX}.

\bibitem[Rigter et~al.(2023)Rigter, Yamada, and Posner]{rigter2023world}
Marc Rigter, Jun Yamada, and Ingmar Posner.
\newblock World models via policy-guided trajectory diffusion, 2023.

\bibitem[Ronneberger et~al.(2015)Ronneberger, Fischer, and Brox]{ronneberger2015u}
Olaf Ronneberger, Philipp Fischer, and Thomas Brox.
\newblock U-net: Convolutional networks for biomedical image segmentation.
\newblock In \emph{Medical Image Computing and Computer-Assisted Intervention--MICCAI 2015: 18th International Conference, Munich, Germany, October 5-9, 2015, Proceedings, Part III 18}, pp.\  234--241. Springer, 2015.

\bibitem[Sims et~al.(2024)Sims, Lu, and Teh]{sims2024edgeofreach}
Anya Sims, Cong Lu, and Yee~Whye Teh.
\newblock The edge-of-reach problem in offline model-based reinforcement learning, 2024.

\bibitem[Sohl-Dickstein et~al.(2015)Sohl-Dickstein, Weiss, Maheswaranathan, and Ganguli]{pmlr-v37-sohl-dickstein15}
Jascha Sohl-Dickstein, Eric Weiss, Niru Maheswaranathan, and Surya Ganguli.
\newblock Deep unsupervised learning using nonequilibrium thermodynamics.
\newblock In Francis Bach and David Blei (eds.), \emph{Proceedings of the 32nd International Conference on Machine Learning}, volume~37 of \emph{Proceedings of Machine Learning Research}, pp.\  2256--2265, Lille, France, 07--09 Jul 2015. PMLR.
\newblock URL \url{https://proceedings.mlr.press/v37/sohl-dickstein15.html}.

\bibitem[Sun(2023)]{offinerlkit}
Yihao Sun.
\newblock Offlinerl-kit: An elegant pytorch offline reinforcement learning library.
\newblock \url{https://github.com/yihaosun1124/OfflineRL-Kit}, 2023.

\bibitem[Sutton \& Barto(2018)Sutton and Barto]{Sutton1998}
Richard~S. Sutton and Andrew~G. Barto.
\newblock \emph{Reinforcement Learning: An Introduction}.
\newblock The MIT Press, second edition, 2018.
\newblock URL \url{http://incompleteideas.net/book/the-book-2nd.html}.

\bibitem[Todorov et~al.(2012)Todorov, Erez, and Tassa]{mujoco}
Emanuel Todorov, Tom Erez, and Yuval Tassa.
\newblock Mujoco: A physics engine for model-based control.
\newblock \emph{IEEE}, pp.\  5026--5033, 2012.
\newblock URL \url{http://dblp.uni-trier.de/db/conf/iros/iros2012.html#TodorovET12}.

\bibitem[Wang et~al.(2023)Wang, Hunt, and Zhou]{wang2023diffusion}
Zhendong Wang, Jonathan~J Hunt, and Mingyuan Zhou.
\newblock Diffusion policies as an expressive policy class for offline reinforcement learning, 2023.

\bibitem[Yu et~al.(2020)Yu, Thomas, Yu, Ermon, Zou, Levine, Finn, and Ma]{mopo}
Tianhe Yu, Garrett Thomas, Lantao Yu, Stefano Ermon, James~Y Zou, Sergey Levine, Chelsea Finn, and Tengyu Ma.
\newblock Mopo: Model-based offline policy optimization.
\newblock In H.~Larochelle, M.~Ranzato, R.~Hadsell, M.F. Balcan, and H.~Lin (eds.), \emph{Advances in Neural Information Processing Systems}, volume~33, pp.\  14129--14142. Curran Associates, Inc., 2020.
\newblock URL \url{https://proceedings.neurips.cc/paper/2020/file/a322852ce0df73e204b7e67cbbef0d0a-Paper.pdf}.

\bibitem[Zhang et~al.(2023)Zhang, Xiong, Yang, Casas, Hu, and Urtasun]{zhang2023learning}
Lunjun Zhang, Yuwen Xiong, Ze~Yang, Sergio Casas, Rui Hu, and Raquel Urtasun.
\newblock Learning unsupervised world models for autonomous driving via discrete diffusion.
\newblock \emph{arXiv preprint arXiv:2311.01017}, 2023.

\end{thebibliography}
\bibliographystyle{rlc}


\clearpage
\appendix

\section*{\LARGE \centering Appendix}

\section{Hyperparameters\label{sec:all-hypers}}
We open-source our implementation at \href{https://github.com/EmptyJackson/policy-guided-diffusion}{https://github.com/EmptyJackson/policy-guided-diffusion}.

\subsection{Diffusion Model}
For the diffusion model, we used a U-Net architecture~\citep{ronneberger2015u} with hyperparameters outlined in \autoref{tab:unet_hyp}.
We transformed the trajectory by stacking the observation, action, reward, and done flags for each transition, before performing 1D convolution across the sequence of transitions.

\begin{table}[h!]
    \centering
    \caption{U-Net hyperparameters}
    \begin{tabular}{@{}rl@{}}
        \toprule
        \textbf{Hyperparameter} & \textbf{Value} \\
        \midrule
        Trajectory length & 16 \\
        Kernel size & 3 \\
        Features & 1024 \\
        U-Net blocks & 3 \\
        \cmidrule{2-2}
        Batch size & 16 \\
        Dataset epochs & 250 \\
        \cmidrule{2-2}
        Optimizer & Adam \\
        Learning rate & $2 \times 10^{-3}$ \\
        LR schedule & Cosine decay \\
        \bottomrule
    \end{tabular}
    \label{tab:unet_hyp}
\end{table}

\subsection{Diffusion Sampling}
We use EDM~\citep{karras2022elucidating} for diffusion sampling, retaining many of the default hyperparameters from \citet{lu2023synthetic} (\autoref{tab:edm_hyp}).
We tuned the number of diffusion timesteps, finding diminishing improvement in dynamics error beyond 256 timesteps.
\begin{table}[h!]
    \centering
    \caption{EDM hyperparameters}
    \begin{tabular}{@{}rl@{}}
        \toprule
        \textbf{Hyperparameter} & \textbf{Value} \\
        \midrule
        Diffusion timesteps & 256 \\
        $\text{S}_{\text{churn}}$ & 80 \\
        $\text{S}_{\text{noise}}$  & $1.003$ \\
        $\text{S}_{\text{tmax}}$  & 50 \\
        $\text{S}_{\text{tmin}}$  & $0.05$ \\
        $\sigma_{\text{max}}$  & 80 \\
        $\sigma_{\text{min}}$  & $0.002$ \\

        \bottomrule
    \end{tabular}
    \label{tab:edm_hyp}
\end{table}

\clearpage
\section{Noised Target Distribution}
\label{sec:noised-target}
To model the target distribution with diffusion, we require the noise-conditioned score function $\nabla_{\bm{\hat{\tau}}} \log \ptarg(\bm{\hat{\tau}};\sigma)$ for the target distribution.
However, since we do not have access to samples from $p_\text{target}(\bm{\hat{\tau}};\sigma)$, one might wish to apply a factorization of the target distribution, such as
\begin{align}
    \ptarg(\bm{\tau}) &= \poff(\bm{\tau}) \prod_{t=0}^{H-1} \frac{\pi_{\text{target}}(a|s)}{\pi_{\text{off}}(a|s)}\text{,}
\end{align}
before modeling its terms separately.
However, by applying independent Gaussian noise to each of the elements in $\bm{\hat{\tau}}$, we lose conditional independence between contiguous states and actions---i.e., $\ptarg(\hat{a}_t | \bm{\hat{\tau}}\! \setminus\! \hat{a}_t;\sigma) \neq \ptarg(\hat{a}_t | \hat{s}_t;\sigma)$---preventing us from applying an equivalent factorization.
Due to this, we must approximate $\nabla_{\bm{\hat{\tau}}} \log \ptarg(\bm{\hat{\tau}};\sigma)$ directly, as we propose in \autoref{sec:equi-policy}.

\section{Behavior-Regularized Target Distribution\label{sec:dist-motiv}}
Intuitively, the behavior-regularized target distribution transforms the target distribution by increasing the likelihood of actions under the behavior policy.
As is typical in offline RL~\citep{kumar2020conservative, fujimoto2021minimalist, bcq}, regularizing the policy towards the behavior distribution is required in order to avoid out-of-sample states and consequently minimize value overestimation.
Rather than regularizing the policy, PGD shifts this regularization to the data generation process, which helps our guided samples remain in-distribution with respect to the diffusion model, and thus less susceptible to model error.

Moreover, we note that this type of regularization is not immediately available for prior autoregressive world models, and thus they typically penalize reward by dynamics error~\citep{mopo, kidambi2020morel, lu2022revisiting} in an ad-hoc fashion in order to avoid model exploitation.
In contrast, PGD presents a \emph{natural mechanism for behavioral regularization during data generation}, making offline policy optimization without regularization a promising path for future work.

\section{Agent Training with Policy-Guided Diffusion}
In \autoref{alg:training}, we present pseudocode for training an agent with synthetic experience generated by PGD. PGD is agnostic to the underlying offline RL algorithm used to train the target policy, making it a drop-in extension to any model-free method.

\begin{algorithm}[htb]
    \flushleft
    \caption{Agent training via {\color{pgd}policy-guided} diffusion.}
    \label{alg:training}
    \begin{algorithmic}[1]
        \State {\bfseries Parameters:} Number of epochs $N_{\text{epochs}}$, steps per epoch $N_{\text{policy}}$
        \State {\bfseries Required:} Diffusion trajectory sampler $\mathcal{F}(\bm{\tau} {\color{pgd}| \pi};\theta, \lambda)$
        \State Initialize policy $\pi_{\phi}$
        \For{epoch $= 0$ {\bfseries to} $N_{\text{epochs}}$}
            \State Generate synthetic dataset $\mathcal{D}_{\text{epoch}} \sim \mathcal{F}(\bm{\tau} {\color{pgd}| \pi_{\phi}};\theta, \lambda)$
            \For{step $= 0$ {\bfseries to} $N_{\text{policy}}$}
                \State Sample mini-batch $\{\bm{\tau}\} \sim \mathcal{D}_{\text{epoch}}$
                \State Update policy $\pi_{\phi}$ on mini-batch $\{\bm{\tau}\}$
            \EndFor
        \EndFor
        \State {\bfseries return} $\pi_{\phi}$
    \end{algorithmic}
\end{algorithm}

\end{document}